\documentclass[a4paper,conference]{IEEEtran}
\usepackage{graphicx}
\usepackage{amsmath}
\usepackage{amssymb}
\usepackage{rotating}
\usepackage{color}
\usepackage{framed}
\usepackage{multirow}
\usepackage[center]{subfigure}
\usepackage{balance}
\usepackage{xspace}
\usepackage{booktabs}
\usepackage{bm}
\usepackage{bbm}
\usepackage{enumitem}
\usepackage{lipsum}
\usepackage[export]{adjustbox}

\def \ie {\emph{i.e.}}
\def \eg {\emph{e.g.}}
\def \etal {\emph{et al.}}
\def \ours {\textbf{Ours}\xspace}

\newcommand{\tit}[1]{\smallbreak\noindent\textbf{#1.}}

\begin{document}

\title{A Novel Attention-based Aggregation Function\\ to Combine Vision and Language}

\author{\IEEEauthorblockN{Matteo Stefanini, Marcella Cornia, Lorenzo Baraldi, Rita Cucchiara}
\IEEEauthorblockA{University of Modena and Reggio Emilia\\
Email: \{name.surname\}@unimore.it}
}

\maketitle

\begin{abstract}
The joint understanding of vision and language has been recently gaining a lot of attention in both the Computer Vision and Natural Language Processing communities, with the emergence of tasks such as image captioning, image-text matching, and visual question answering. As both images and text can be encoded as sets or sequences of elements -- like regions and words -- proper reduction functions are needed to transform a set of encoded elements into a single response, like a classification or similarity score. In this paper, we propose a novel fully-attentive reduction method for vision and language. Specifically, our approach computes a set of scores for each element of each modality employing a novel variant of cross-attention, and performs a learnable and cross-modal reduction, which can be used for both classification and ranking. We test our approach on image-text matching and visual question answering, building fair comparisons with other reduction choices, on both COCO and VQA 2.0 datasets. Experimentally, we demonstrate that our approach leads to a performance increase on both tasks. Further, we conduct ablation studies to validate the role of each component of the approach.
\end{abstract}


\section{Introduction}
As humans we learn to combine vision and language early in life, building connections between visual stimuli and our ability to communicate in a common natural language. The abundance and diversity of daily-created data pose new unparalleled opportunities in the attempt to artificially reproduce this joint visual-semantic understanding.
Recent progress at the intersection of Computer Vision and Natural Language Processing has led to new architectures capable of automatically combining the two modalities, improving the performance of different vision-and-language tasks, such as image captioning~\cite{anderson2018bottom,cornia2020m2}, cross-modal retrieval~\cite{kiros2014unifying,baraldi2018aligning,faghri2018vse,lee2018stacked}, and visual question answering~\cite{andreas2016neural,anderson2018bottom,tan2019lxmert}. All these settings have usually been addressed by using recurrent neural networks that can naturally model the sequential nature of textual data. However, the recent advent of fully attentive mechanisms, in which the recurrent relation is abandoned in favor of the use of self- and cross-attention, has consistently changed the way to deal with visual and textual data, as testified by the success and performance improvements obtained with the Transformer~\cite{vaswani2017attention} and BERT~\cite{devlin2018bert} models. 

\begin{figure}[t]
    \centering
    \includegraphics[width=0.98\linewidth]{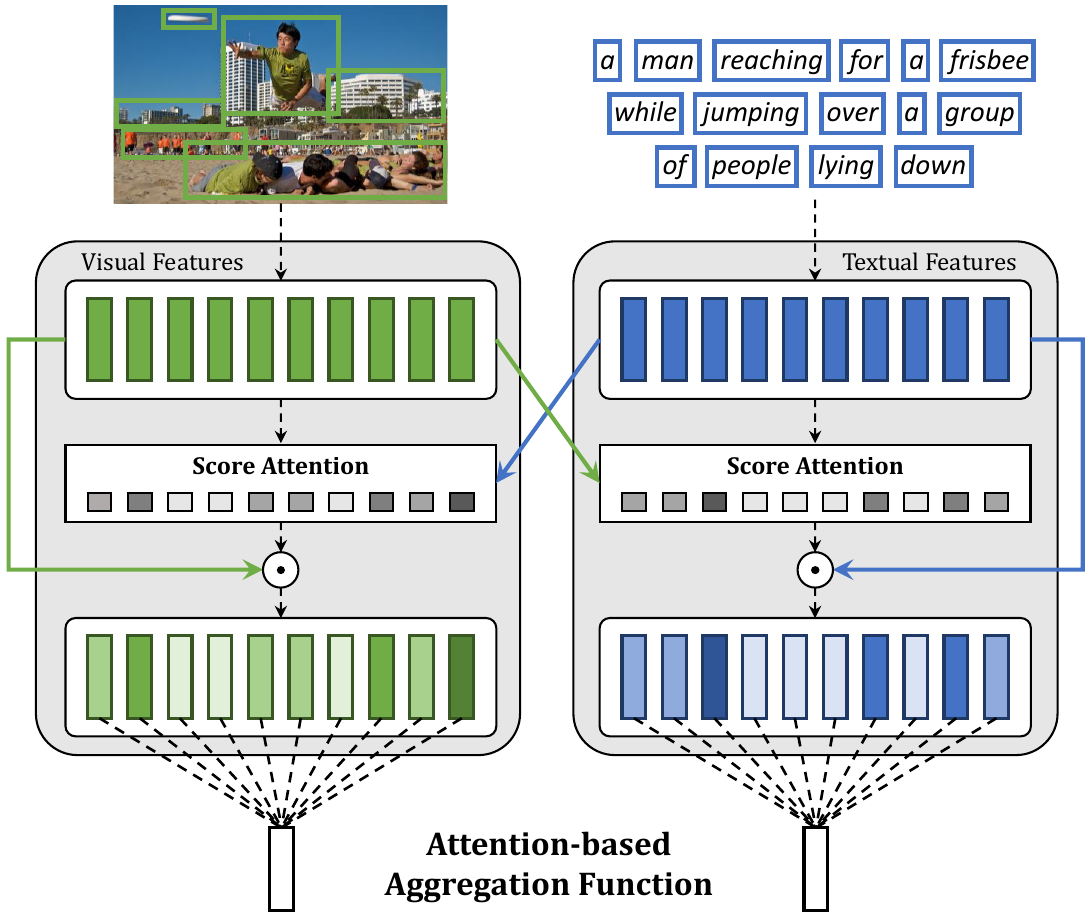}
    \caption{We propose a novel aggregation function for vision-and-language tasks. Given sets of visual and textual inputs, our approach computes a set of \textit{scores} for each modality, using a novel operator based on cross-attention which ensures a learnable reduction based on cross-modal information flow.}
    \label{fig:first_page}
\end{figure}

Nonetheless, the difficulty in tackling these problems is still given by the huge discrepancy between visual-semantic modalities. In this context, recent research efforts have mainly focused on treating images and text as sets or sequences of building elements, such as image regions and sentence words, leading to a better content understanding of both modalities~\cite{anderson2018bottom,lee2018stacked}. While this approach has allowed more fine-grained alignment and richer representation capabilities of visual-semantic concepts, it has also caused a large increase of features that need to be combined together without loosing inter- and intra-modality interactions. 

As such, aggregating features represents one of the crucial steps in visual-semantic tasks in which different information are fused together to obtain a compact and comprehensive representation of both modalities. In this paper, we tackle the problem of aggregating visual-semantic features in an effective and learnable way, and propose a novel aggregation function based on attentive mechanisms that can be successfully applied to different vision-and-language tasks. Our method can be seen as a variant of the cross-attention schema in which a set of scores are learned to aggregate feature vectors coming from image regions and textual words, thus taking into account the cross-modality interactions between elements (Fig.~\ref{fig:first_page}). 

We apply our attention-based aggregation function to cross-modal retrieval and visual question answering: in the first case, the compact representation of visual-semantic data is used to measure the similarity between the input image and the textual sentence, while, in the visual question answering task, it is used to compute a classification score over a set of possible answers for each image-question pair. Experimentally, we test our approach on the COCO dataset for cross-modal retrieval and on the VQA 2.0 dataset for visual question answering, and we demonstrate its effectiveness in both settings with respect to different commonly used aggregation functions.

To summarize, our main contributions are as follows:
\begin{itemize}[noitemsep,topsep=0pt]
\item We introduce a new aggregation method based on attentive mechanisms that learns a compact representation of sets or sequences of feature vectors.
\item We tailor our method to combine vision-and-language data in order to obtain a cross-modal reduction for both classification and ranking objectives. Also, our method can be easily adapted to other tasks requiring an aggregation of elements with minimum changes in the architecture design.
\item We show the effectiveness of our solution when compared to other common reduction operators, demonstrating superior performance in aggregating multi-modal features.
\end{itemize}

\section{Related Work}
In the last few years, several research efforts have been made to improve the performance of cross-modal retrieval and visual question answering methods, resulting in novel architectures~\cite{gu2018look,wang2019position,li2019relation,gao2019multi}, effective training~\cite{kiros2014unifying,faghri2018vse} and pre-training~\cite{tan2019lxmert} strategies, and more powerful representations of images and text~\cite{anderson2018bottom,lee2018stacked,li2019visual}. In the following, we review the most important work related to these two visual-semantic tasks and provide a brief overview of feature aggregation functions used in the deep learning literature. 

\subsection{Cross-modal Retrieval}
The key issue of cross-modal retrieval methods is to measure the visual-semantic similarity between images and textual sentences. Typically, this is achieved by learning a common embedding space where visual and textual data can be projected and compared. One of the first attempt in this direction has been made by Kiros~\etal~\cite{kiros2014unifying} in which a triplet ranking loss is used to maximize the distance between mismatching items and minimize that between matching pairs.

Following this line of work, Faghri~\etal~\cite{faghri2018vse} introduced a simple modification of standard loss functions, based on the use of hard negatives during training, that has been demonstrated to be effective in improving the final performance and widely adopted by several subsequent methods~\cite{engilberge2018finding,gu2018look,lee2018stacked}. Among them, Gu~\etal~\cite{gu2018look} further improved the visual-semantic embedding representations by incorporating generative processes of images and text. Differently, Engilberge~\etal~\cite{engilberge2018finding} proposed a novel approach in which spatial pooling mechanisms are used to embed visual features and localize new concepts from the embedding space.

Recently, strong improvements have been obtained with the stacked cross-attention mechanism proposed by Lee~\etal~\cite{lee2018stacked} in which a latent correspondence between image regions and words of the caption is learned to match images and textual sentences. Wang~\etal~\cite{wang2019position} extended this paradigm by adding the relative position of image regions in the visual encoder, demonstrating better performance. On a similar line, Li~\etal~\cite{li2019visual} introduced a visual-semantic reasoning model based on graph convolutional networks that can generate better visual representations and capture key objects and semantic concepts present on a scene. 

\subsection{Visual Question Answering}
Many different solutions have been proposed to address the VQA task, ranging from Bayesian~\cite{malinowski2014multi} and compositional~\cite{andreas2016neural,hu2017learning} approaches to spatial attention-based methods~\cite{yang2016stacked,anderson2018bottom} and bilinear pooling schemes~\cite{kim2018bilinear}. In the last few years, the use of attention mechanisms has become the leading choice for this task, resulting in new models in which relevance scores over visual and textual features are computed to process only relevant information. Among them, Anderson~\etal~\cite{anderson2018bottom} re-visited the standard attention over a spatial grid of features and proposed to encode images with multiple feature vectors coming from a pre-trained object detector.

After this work, several methods with attention over image regions have been presented~\cite{kim2018bilinear,cadene2019murel,gao2019dynamic,tan2019lxmert,li2019relation}. While Cadene~\etal~\cite{cadene2019murel} proposed a reasoning module to encode the semantic interaction between each visual region and the question, Gao~\etal~\cite{gao2019dynamic} introduced a dynamic fusion framework that integrates inter- and intra-modality information. Differently, Li~\etal~\cite{li2019relation} presented a novel solution based on graph attention networks that considers spatial and semantic relations to enrich image representations. 

Following the advent of fully-attentive mechanisms for sequence modeling tasks like machine translation and language understanding~\cite{vaswani2017attention,devlin2018bert}, different Transformer-based solutions have also been proposed to address multimodal settings~\cite{tan2019lxmert,gao2019multi,cornia2020m2}. In the context of visual question answering, Yu~\etal~\cite{yu2019deep} presented a co-attention module made of a stack of attentive layers based on self-attention, keeping the textual encoder based on recurrent neural networks. Gao~\etal~\cite{gao2019multi}, instead, introduced a novel architecture entirely based on fully-attentive mechanisms that learns cross-modality relationships between latent summarizations of visual regions and questions. On a similar line, Tan~\etal~\cite{tan2019lxmert} proposed a Transformer-based model that has demonstrated improved performance thanks to a pre-training phase on large amounts of image-sentence pairs.

\subsection{Feature Aggregation Methods}
The aggregation of spatial and temporal features is one of the key challenges in deep learning architectures. Different solutions have been proposed and heavily depend on the domain in which applying the aggregation functions (\ie~images or text). While fusing and pooling operations applied over depths, scales, and resolutions constitute fundamental components in visual recognition architectures, the sequential nature of textual data requires different strategies to reduce feature dimensionality.

Regarding the visual domain, with the first strategies adopted in early popular deep learning models~\cite{krizhevsky2012imagenet,simonyan2014very,szegedy2015going}, the architecture design has moved in last few years to deeper and wider networks~\cite{he2016deep,xie2017aggregated,huang2017densely} incorporating bottlenecks and connectivity novelties like skipping, gating, and aggregating mechanisms. While going deeper, \ie~aggregating across channels and depths, improves the semantic recognition accuracy, spatial fusion, \ie~aggregating across scales and resolutions, is needed to achieve a better localization capability. In this context, feature pyramid networks~\cite{lin2017feature} are the predominant approach, making use of top-down and lateral connections between feature hierarchical levels.

On a different note, data with a sequential nature such as textual sentences require different solutions to take into account the temporal dependencies between elements. In this setting, the use of recurrent neural networks has remained the most commonly used strategy, where hidden representations, learned through memory and gating mechanisms, are adopted as global encoding of a sequence of feature vectors.

Recently, with the advent of fully attentive architectures~\cite{vaswani2017attention} that overcame limitations of recurrent networks, novel solutions based their global understanding of sequences through the addition of a special CLS token at the beginning of each sequence~\cite{devlin2018bert,tan2019lxmert}. Thanks to the use of attention that models inter- and intra-modality connections, this CLS token can learn a compact representation of an input sequence for general classification purposes. Additionally, similar efforts have been made on the encoding of textual sentences, where again mean and max pooling or CLS token have remained the predominant aggregation approaches~\cite{humeau2019poly,reimers2019sentence}. 

Differently from previous works, we propose a novel aggregation method based on attentive mechanisms that can reduce in a learnable way a set or a sequence of features coming from either the visual or textual domain.

\begin{figure*}[t]
    \centering
    \includegraphics[width=\linewidth]{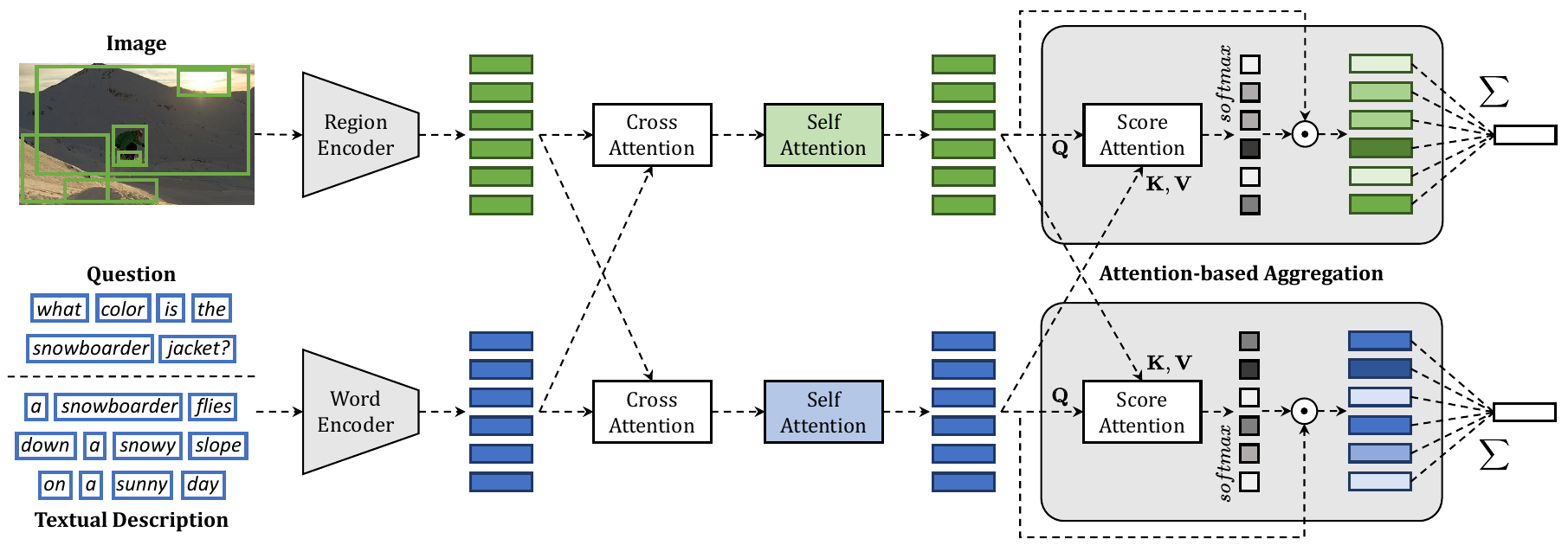}
    \caption{Our architecture for cross-modal feature extraction and matching. After a cross-modal feature extraction stage, the proposed attention-based aggregation function aligns and reduces vectors from both modalities into compact and cross-modal representations.}
    \label{fig:score_attention}
\end{figure*}

\section{Proposed Method}
The popular scaled dot-product attention mechanism operates on a set of input vectors and generates a new set of vectors, where each one has been updated with relevant information coming from the others. 
This has been proved largely effective in sequence to sequence tasks such as natural language understanding and machine translation~\cite{vaswani2017attention,devlin2018bert}. However, visual-semantic tasks such as visual question answering and cross-modal retrieval deal with multi-modal input sequences and require alignment between modalities and global perspectives in order to reach a classification or similarity output.

To this end, we propose a novel attention-based aggregation function that learns to align and combine two sets of features into a global and compact representation based on the cross-domain connections between modalities. In the simplest case, the two sets of features will be regions from an input image and word features from a natural language sentence.
In a nutshell, our approach leverages dot-product attention to compute cross-modal scores for each element of the two feature sets. The weights are then used to take a weighted sum of the input feature vectors, thus reducing the two sets into a pair of vectors which can be used for classification or ranking.

In the following, we firstly present our attention-based reduction method. With the aim of testing the operator on both image-text matching and visual question answering, we then introduce a general architecture for both tasks, where features from multi-modal inputs are extracted and combined. In the last section, we discuss the final stages of the architecture and the training choices.

\subsection{Attention-based Aggregation Function}
Motivated by the need of leveraging the information contained in sequence of vectors and at the same time to compare multi-modal information, our aggregation function is based on the scaled dot-product attention mechanism~\cite{vaswani2017attention}.

To recall what attention is, given three sets of vectors, \ie~queries $\bm{Q}$, keys $\bm{K}$ and values $\bm{V}$, scaled dot-product attention computes a weighted sum of the value vectors according to a similarity distribution between query and key vectors. This is usually done in a multi-head fashion, so that for each head $h$ the attention operator is defined as
\begin{align}
\mathsf{Attention}_h(\bm{Q}_h, \bm{K}_h, \bm{V}_h)=\operatorname{softmax}\left(\frac{\bm{Q}_h \bm{K}_h^{T}}{\sqrt{d}}\right) \bm{V}_h,
\label{eq:attention}
\end{align}
where $\bm{Q}_h$ is a matrix of $n_q$ query vectors, $\bm{K}_h$ and $\bm{V}_h$ both contain $n_k$ keys and values, and $d$ is the dimensionality of queries and keys, used as a scaling factor.

In the case of self-attention, queries, keys and values are obtained for each head as linear projections of the same input vectors belonging to a single modality, while in cross-attention, queries are a projection of one modality vectors and keys and values are projections of the other modality vectors.
Inspired by cross-attention, we define a \textit{Score Attention} operator which computes a relevance score for each element of the query sequence, considering its relationships with keys and values coming from the other modality.

Formally, given the set of query, key and value vectors from all heads, our Score dot-product attention can be formulated as 
\begin{align}
\mathsf{ScoreAttn}(\mathcal{Q}, \mathcal{K}, \mathcal{V})=\operatorname{fc}\left(\left[\operatorname{softmax}\left(\frac{\bm{Q}_h \bm{K}_h^{T}}{\sqrt{d}}\right) \bm{V}_h\right]_h\right),
\label{eq:scoreattention}
\end{align}
where $\mathcal{Q}$, $\mathcal{K}$, $\mathcal{V}$ indicate the set of queries, keys and values for the different heads, $[...]_h$ indicates the concatenation of the outputs of all heads and $\operatorname{fc}$ is a linear projection that outputs a single scalar score for each input query.

In order to learn complex interactions between modalities and therefore to guide the reduction process based on the cross-domain relations, Score Attention is applied on  queries from one modality and keys and values from the other modality.
Therefore, given a set of input vectors $\bm{X}$ coming from one modality (\textit{e.g.} regions of an image) and a set of input vectors $\bm{Z}$ coming from the other modality (\textit{e.g.} words of a text), we obtain a final condensed representation for $\bm{X}$ conditioned on $\bm{Z}$ as a weighted sum of its vectors using the scores provided by the Score Attention operator, \ie 
\begin{align}
\bm{Y}(\bm{X}, \bm{Z})=\sum_{i=0}^{n_q} \bm{S}_i(\bm{X}, \bm{Z}) \cdot \bm{X}_i
\label{eq:output_vector}
\end{align}
\begin{align}
\bm{S}(\bm{X}, \bm{Z}) = \operatorname{softmax}\left(\mathsf{ScoreAttn}(\mathcal{Q}, \mathcal{K}, \mathcal{V})\right),
\label{eq:scores}
\end{align}
where queries $\mathcal{Q}$ are obtained as projections of $\bm{X}$, while keys and values $\mathcal{K}$, $\mathcal{V}$ are obtained from $\bm{Z}$. The softmax is applied over the $n_q$ scores returned by the Score Attention operator. Conversely, the same applies to the reduction of the other modality $\bm{Z}$ by considering $\bm{Z}$ as query sequence and $\bm{X}$ as key and value ones.

As it can be seen from Eq.~\eqref{eq:scoreattention} and \eqref{eq:scores}, our Score Attention operator can be thought as a cross-attention that, instead of yielding a sequence of vectors, computes a sequence of scores conditioned on keys and values from the other modality. 
Therefore, the final compressed representation for each modality can capture a global perspective of the input, focusing on elements that show higher importance with respect to the cross-domain interactions.

Noticeably, this aggregation function can be executed multiple times in parallel with different query, key and value projections, thus yielding more than one output vector. This in principle can foster a more disentangled representation, in which different output vectors refer to different global aspects of the same input features.
We therefore test our method with different number of compressed vectors, and we refer later to this hyper-parameter with $k$.
Whenever the number of vectors is more than one, we average their contributions with a non-learnable reduction operator. More details on this can be found in the Implementation Details section.

\subsection{Visual-Semantic Model} \label{sec:pipeline}
To test our aggregation operator, we devise a general architecture for cross-modal feature extraction and matching, with the aim of tackling different tasks with the same common pipeline. Specifically, the architecture is tested on both image-text retrieval and visual question answering.
Given input regions from an image, and words from a textual description, we adopt a bi-directional GRU as text encoder, retaining for each word the average embedding between the forward hidden state and the backward hidden state. On the visual side, instead, we apply a linear projection to the features of image regions.

Following recent progress in fully-attentive models and cross-modality interaction \cite{vaswani2017attention, tan2019lxmert}, after this encoding stage we propagate visual and textual features with a cross-attention operation, followed by a self-attention for each modality. On top of this, two instances of the aggregation operator are applied, one for each modality, thus obtaining one global vector for each modality. A summary of the overall architecture is reported in Fig.~\ref{fig:score_attention}.

\subsection{Training}
The last stage of the model and the training objectives depend on the specific task. In the following we report the main differences.

\tit{Visual Question Answering}
After applying the aggregation operator, the two vector representations are concatenated and fed to a fully connected layer which is in charge of predicting the final answer class.
Additionally, in the case of VQA, we add a position-wise feed-forward layer between the reduction operator and the final concatenation for class prediction.

During the training phase, we employ the binary cross-entropy loss in a multi-label fashion, \ie~applying it independently for all classes. For fairness of comparison, we do not make use of any data augmentation strategy and do not employ any external data source like part of the VQA literature does.

\tit{Cross-modal Retrieval} 
In the case of image-text matching, instead, the compressed vectors given by the application of the aggregation operator are compared with a cosine similarity to measure their similarity score.
During training, we adopt an hinge-based triplet ranking loss, which is the most common ranking objective in the retrieval literature. Following Faghri~\etal~\cite{faghri2018vse}, we only backpropagate the loss obtained on the hardest negatives found in the mini-batch. Given image and sentence pairs $(I, T)$, our final loss with margin $\alpha$ is thus defined as
\begin{align*}
L_{hard}(I, T)=\max _{\hat{T}}\left[\alpha -S(I, T) + S(I, \hat{T})\right]_{+} \\ + \max _{\hat{I}}\left[\alpha -S(I, T) + S(\hat{I}, T) \right]_{+},
\end{align*}
where $S$ indicates the cosine similarity, $[x]_{+} = \max(x,0)$, $\hat{T}$ is the hardest negative sentence and $\hat{I}$ is the hardest negative image.

\section{Experimental Evaluation}
In this section, we report the results on the two considered visual-semantic tasks (\ie~visual question answering and cross-modal retrieval) by comparing our attention-based aggregation function with respect to different baselines. First, we provide implementation details and introduce the datasets used in our experiments.

\subsection{Datasets}
To validate the effectiveness of our solution, we employ two of the most widely used datasets containing visual-semantic data. In particular, we carry out the experiments on the VQA 2.0~\cite{goyal2017making} and COCO~\cite{lin2014microsoft} datasets to address visual question answering and cross-modal retrieval, respectively.

\tit{COCO} The dataset contains more than $120\,000$ images, each of them annotated with $5$ different textual descriptions. We follow the splits provided by Karpathy~\etal~\cite{karpathy2015deep}, where $5\,000$ images are used for validation, $5\,000$ for testing and the rest for training. Following the standard evaluation protocol~\cite{faghri2018vse}, retrieval results on this dataset are reported by averaging over $5$ folds of $1\,000$ test images each.

\tit{VQA 2.0} The dataset is composed of images coming from the COCO dataset and are divided in training, validation, and test according to the official splits. For each image, three questions are provided on average. These questions are divided into three different categories: \texttt{Yes/No}, \texttt{Number}, and \texttt{Others}. Each image-question pair is annotated with $10$ answers collected by human annotators, and the most frequent answer is selected as the correct one. We report experimental results on the validation and test-dev sets of this dataset, only using the training split to train our model. Differently from standard literature that uses additional training data coming from different datasets, we only focus on image-question-answer triplets from this dataset.

\subsection{Implementation Details}
To encode image regions, we employ the Faster R-CNN model finetuned on the Visual Genome dataset~\cite{krishnavisualgenome,anderson2018bottom}, obtaining a $2048$-dimensional feature vector for each region. We reduce the dimensionality of region feature vectors by feeding them to a fully connected layer with a size of $512$. For each image, we select the top $36$ regions with the highest class detection confidence score. As mentioned, to encode word vectors, we use a bi-directional GRU with a single layer using either learned or pre-trained word embeddings to represent words of the sentence. We set the hidden size of the GRU layer to $512$.

Following the standard implementation~\cite{vaswani2017attention}, each scaled dot-product attention also includes a dropout, a residual connection, and a layer normalization. We set the dimensionality $d$ of each layer to $512$, the number of heads in both scaled dot-product and score attention to $8$, and the dropout keep probability to $0.9$. In all our experiments, we use Adam~\cite{kingma2015adam} as optimizer and a batch size equal to $64$. 

\tit{Visual Question Answering}
For VQA models, we set the initial learning rate to $0.0005$ decreased by a factor of $10$ every $10$ epochs. To represent words, we use and finetune the pre-trained GloVe word embeddings~\cite{pennington2014glove} with a word dimensionality equal to $300$. We set the maximum length of input questions to $14$, padding the shorter ones. For the additional position-wise feed forward layer used in VQA models, we set the hidden size to the same dimensionality $d$ of attention layers. When we use a number of compressed vectors $k$ larger than $1$, we average the $k$ vectors of each modality to obtain a single compact representation for both image regions and words.

Following a common practice in the VQA task~\cite{anderson2018bottom}, the set of candidate answers is limited to correct answers in the training set that appear more than $8$ times, resulting in an output vocabulary size equal to $3\,129$.

\begin{table*}[t]
    \centering
    \caption{Accuracy results on VQA 2.0 dataset. The results are reported on the validation and test-dev splits. All models are trained only on the VQA 2.0 training split.}
    \label{tab:vqa_results}
    \setlength{\tabcolsep}{.85em}
    \begin{tabular}{lcccccccccc}
    \toprule
    & & \multicolumn{4}{c}{\textbf{Validation}} & & \multicolumn{4}{c}{\textbf{Test-Dev}} \\
    \cmidrule{3-6} \cmidrule{8-11}
    \textbf{Aggregation Function} & & All & Yes/No & Number & Others & & All & Yes/No & Number & Others \\
    \midrule
    Mean Pooling        & & 54.87 & 71.50 & 37.93 & 46.69 & & 56.05 & 71.00 & 38.88 & 47.19 \\
    Max Pooling         & & 56.73 & 75.68 & 37.64 & 47.37 & & 57.95 & 75.14 & 38.48 & 47.69 \\
    LogSumExp Pooling         & & 54.61 & 70.94 & 38.27 & 46.53 & & 55.68 & 70.36 & 38.72 & 47.00 \\
    1D Convolution     & & 56.87 & 72.35 & 39.18 & 49.79 & & 57.79 & 71.71 & 39.97 & 49.96 \\
    CLS Token           & & 58.31 & 74.29 & 39.89 & 51.03 & & 59.40 & 74.26 & 40.31 & 51.07 \\
    \midrule
   \ours $(k=1)$   & & 60.73 & 77.68 & 41.86 & \textbf{52.84} & & 62.05 & 77.84 & 42.47 & \textbf{53.03} \\
   \ours $(k=2)$   & & 60.76 & 78.06 & 42.32 & 52.48 & & 62.06 & 78.26 & 42.62 & 52.66 \\
    \ours $(k=3)$   & &  60.50 & 77.82 & 41.56 & 52.33 & & 61.80 & 78.22 & 41.69 & 52.35 \\
    \ours $(k=5)$   & & \textbf{60.99} & \textbf{78.62} & 42.53 & 52.46 & & 62.17 & 78.52 & 42.27 & 52.74 \\
    \ours $(k=7)$   & & 60.95 & 78.40 & \textbf{42.65} & 52.53 & & \textbf{62.43} & \textbf{78.75} & \textbf{43.33} & 52.83 \\
    \ours $(k=10)$   & & 59.94 & 77.30 & 40.82 & 51.80 & & 61.16 & 77.39 & 40.69 & 51.97 \\
    \bottomrule
    \end{tabular}
\end{table*}

\tit{Cross-modal Retrieval}
We set the initial learning rate to $0.00007$ decayed by a factor of $10$ every $10$ epochs, and the margin $\alpha$ of the triplet loss function to $0.2$. Also, we clip the $2$-norm of vectorized gradients to $2.0$. To encode words, we use one-hot vectors and linearly project them with a learnable embedding matrix to the word dimensionality of $300$. To create the word vocabulary, we take into account only the words that appear at least $5$ times in the training and validation sets.

In our attention-based aggregation function, when the number of compressed vectors $k$ is larger than $1$, we compute a pair-wise cosine similarity between each pair of compressed vectors coming from the two modalities, and we average the resulting $k$ similarity scores. Intuitively, each aggregation module learns to extract and compare different relevant information, specializing each vector to distinct semantic meaning.

\subsection{Baselines}
To evaluate the proposed method, we compare our results with respect to five different aggregation functions, namely mean pooling, max pooling, log-sum-exp pooling, 1D convolution, and CLS token. For all baselines, we employ the pipeline defined in Sec.~\ref{sec:pipeline}, and the same hyper-parameters and implementation choices used for our architecture.

\tit{Mean Pooling} The mean aggregation function is one of the most common approaches for feature reduction and refers to the global average pooling between each vector of the input sequence. Since input sequences may have different lengths, in our experiments the mean pooling operation is computed using only the valid elements of the sequence.

\tit{Max Pooling} Similarly to the mean operation, the max pooling is another commonly used strategy to reduce feature dimensionality and selects the maximum activation in the feature maps. In our setting, we apply max pooling to the sequence dimension, thus obtaining a single summarized vector for each input sequence.

\tit{LogSumExp Pooling} It can be considered as a smooth approximation of the maximum function and is defined as the logarithm of the sum of the argument exponentials. We apply this operation along the feature dimension thus condensing the most important features for each vector of the sequence.

\tit{1D Convolution} Convolution is the fundamental operation of CNNs and works well for identifying patterns in data. We test 1D convolutions applied to the sequence dimension to obtain a compact and aggregated representation of the whole set of vectors. In our experiments, we set the kernel size equal to the input sequence length. 

\tit{CLS Token} Following the recent progress of pre-training strategies and cross-modality matching~\cite{devlin2018bert}, we also consider the integration of a special CLS token at the beginning of each input sequence. Thanks to the cross- and self-attention operations, the CLS token can be used as a final compact representation of the entire sequence. We add a CLS token for each modality and use them in last stage of the pipeline according to the specific task.

\subsection{Visual Question Answering Results}
Experimental results for the VQA task are shown in Table~\ref{tab:vqa_results} by comparing our aggregation function with respect to the aforementioned baselines. For each method, we report the accuracy on all image-question pairs of the considered splits and the accuracy values on the three question categories of the VQA 2.0 dataset (\ie~\texttt{Yes/No}, \texttt{Number}, and \texttt{Others}).

As it can be seen, our method surpasses all other aggregation functions by a significant margin on both validation and test-dev splits. With respect to the CLS token, which is the top performing baseline in this task, our solution achieves an improvement of $2.68$\% and $3.03$\% in terms of overall accuracy on the validation and test-dev splits, respectively.

Additionally, we test our attention-based aggregation method by using a different number of $k$ compressed vectors and different word embedding strategies. In the bottom section of Table~\ref{tab:vqa_results}, we report the accuracy results by varying the number of compressed vectors. As it can be noticed, the model with $1$ vector reaches good results surpassing all other baselines. Nevertheless, higher performances can be achieved with $5$ and $7$ compressed vectors suggesting that a correct answer can be positively influenced by capturing different aspects of the input features. Above a certain numbers of $k$ vectors, we instead observe a degradation of the performance, as demonstrated by the results with $10$ vectors. This can be explained by the greater complexity of the model that undermines the benefits of learning different global vectors.

\begin{table}[t]
    \centering
    \caption{Comparison between different word embedding strategies on VQA 2.0 validation set.}
    \label{tab:ablatives_word_emb}
    \setlength{\tabcolsep}{0.38em}
    \begin{tabular}{lcccccc}
    \toprule
    \textbf{Aggregation Func.} & \textbf{Word Emb.} & & All & Yes/No & Number & Others \\
    \midrule
    \ours $(k=5)$   & Learned & & 59.29 & 77.24 & 40.29 & 50.66 \\
    \ours $(k=5)$    & GloVe & & 60.98 & 78.51 & 42.20 & \textbf{52.61} \\
    \ours $(k=5)$     & GloVe Finetuned & & \textbf{60.99} & \textbf{78.62} & \textbf{42.53} & 52.46 \\
    \midrule
    \ours $(k=7)$  & Learned & & 59.23 & 76.98 & 40.02 & 50.80 \\
    \ours $(k=7)$  & GloVe & & \textbf{61.13} & \textbf{79.13} & 42.13 & 52.47 \\
    \ours $(k=7)$  & GloVe Finetuned & & 60.95 & 78.40 & \textbf{42.65} & \textbf{52.53} \\
    \bottomrule
    \end{tabular}
\end{table}

\begin{figure*}[t]
    \centering
    \includegraphics[width=0.98\linewidth]{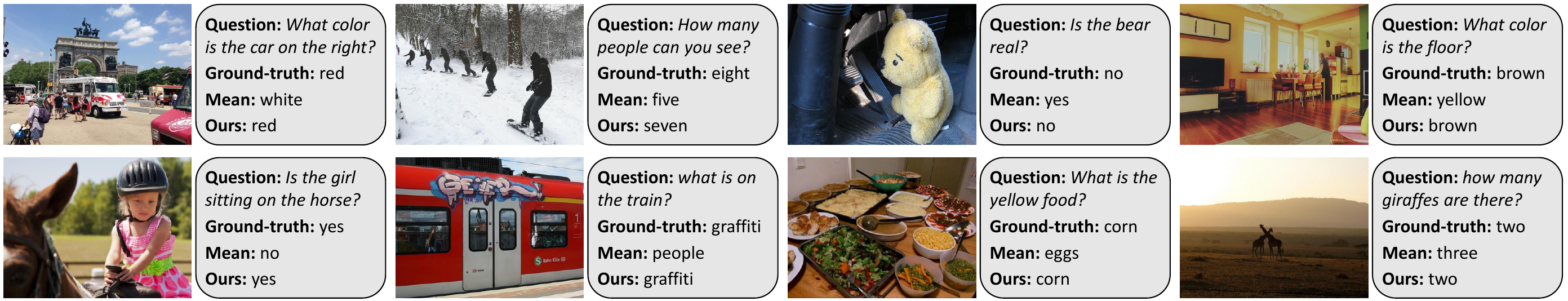}
    \caption{Qualitative results on VQA 2.0 validation set. For each image, we report a sample question, the ground-truth answer, and the corresponding answers predicted by our aggregation function and by the mean pooling operation.}
    \label{fig:vqa_results}
\end{figure*}

In Table~\ref{tab:ablatives_word_emb}, we show the performance on the VQA 2.0 validation set when using different word embedding strategies. In particular, we compare the results by employing learnable word embeddings and pre-trained GloVe vectors, either fixed or finetuned during training. In our experiments, the GloVe word embeddings lead to an improvement of the final accuracy results using both $5$ and $7$ compressed vectors. The performance gap between fixed and finetuned GloVe vectors is not very large, but a slight improvement is given when using the finetuned version. For this reason, all experiments are carried out by using the GloVe vectors finetuned during training. On the contrary, learning word embeddings from scratch brings to lower performances in all settings.

\tit{Qualitative Results} Some sample results on the VQA 2.0 validation set are reported in Fig.~\ref{fig:vqa_results}. For each image, we show the corresponding question, the ground-truth correct answer and the answers predicted by our attention-based aggregation function and the mean pooling operation. The results demonstrate the effectiveness of our strategy also from a qualitative point of view and confirm better performance than one of the most widely used solution to aggregate features. Our method is able to correctly identify the color of the objects contained in the question and count the number of instances of a given entity. Also, it can accurately answer either simple (\eg~\texttt{Yes/No}) or more complex questions that require a complete understanding of the scene.

\subsection{Cross-modal Retrieval Results}
Table~\ref{tab:vse_results} shows the results for the cross-modal retrieval task on the COCO test set. For both text and image retrieval, we report the results in terms of recall@$K$ (with $K=1,5,10$) which measures the portion of query images or query captions for which at least one correct result is found among the top-$K$ retrieved elements. Also in this setting, we compare our aggregation function with respect to the previously defined baselines and we analyze the performance by varying the number of compressed vectors used to aggregate input sequences.

As it can be seen, our attention-based aggregation achieves the best results among all considered aggregation functions on both text and image retrieval. Also in this case, the CLS token results to be the top performing baseline according to all evaluation metrics, confirming the importance of using inter- and intra-modality interactions to reduce feature dimensionality. 

Differently from the VQA task in which the best results are obtained with $5$ and $7$ compressed vectors, the best performances are instead achieved with a lower number of vectors (\ie~$2$ and $3$), as shown in the bottom section of Table~\ref{tab:vse_results}. In this setting, we do not find beneficial the use of GloVe word vectors and all results are thus obtained by learning word embeddings during training. This suggests that the large amount of textual data contained in the COCO dataset compared to that available for the VQA task can lead to specific and more suited word embedding representations.

\begin{figure*}[t]
 \centering
 \subfigure[Text Retrieval]{
 \includegraphics[width=0.98\linewidth]{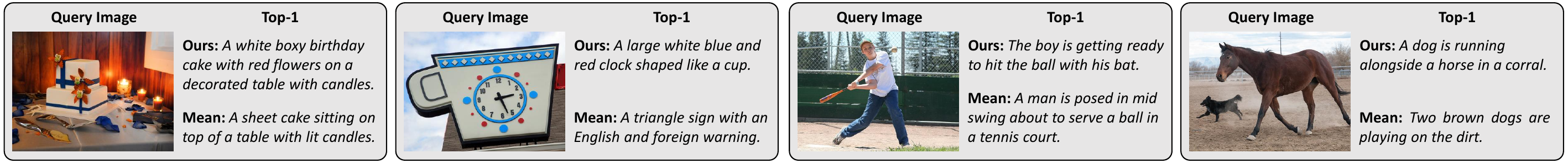}\label{fig:vse_i2t}}
 \subfigure[Image Retrieval]{
 \includegraphics[width=0.98\linewidth]{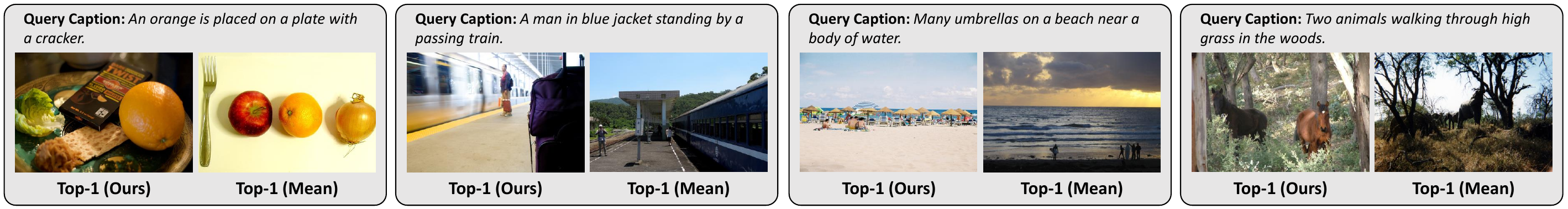}\label{fig:vse_t2i}}
 \caption{Qualitative results for text and image retrieval. For each sample, we report the top-1 result retrieved by our aggregation function and by the mean pooling operation.}
 \label{fig:vse_results}
 \end{figure*}

\tit{Qualitative Results} Finally, we show some sample results for text and image retrieval in Fig.~\ref{fig:vse_i2t} and~\ref{fig:vse_t2i}, respectively. Also in this case, we compare our results with those obtained by using the mean pooling aggregation function. As it can be seen, these qualitative results further confirm the effectiveness of our solution leading to increased and more accurate performance on both settings.

\section{Conclusion}
Aggregating features has always played a critical role in deep learning architectures. In this paper, we proposed a novel aggregation function based on a variant of the cross-attention mechanism, that reduces sets or sequences of elements into a single compact representation in a learnable fashion. We specifically tailored our method for visual-semantic tasks such as visual question answering and cross-modal retrieval where images and text are combined to obtain a classification or a ranking score, respectively. Experimental results demonstrated that our approach achieves better performances when compared to other commonly used reduction functions on both considered tasks.
Further, we showed that our method can be applied with minimum changes in the overall design to different settings where any form of element reduction is required.

\begin{table}[t]
\caption{Cross-modal retrieval results on Microsoft COCO 1K test set.} \label{tab:vse_results}
\centering
\setlength{\tabcolsep}{.45em}
\begin{tabular}{lcccccccc}
\toprule
& & \multicolumn{3}{c}{\textbf{Text Retrieval}} & & \multicolumn{3}{c}{\textbf{Image Retrieval}} \\
\cmidrule{3-5} \cmidrule{7-9}
\textbf{Aggregation Function} & & R@1 & R@5 & R@10 & & R@1 & R@5 & R@10 \\
\midrule
Mean Pooling     & & 69.66 & 93.12 & 97.64 & & 50.42 & 82.27 & 90.83  \\
Max Pooling     & & 69.04 & 92.68 & 96.98 & & 51.20 & 83.27 & 91.52  \\
LogSumExp Pooling    & & 64.20 & 91.52 & 96.84 & & 47.22 & 82.26 & 91.23  \\
1D Convolution          & & 65.66 & 91.86 & 96.58 & & 49.25 & 81.43 & 90.42  \\
CLS Token       & & 70.30 & 93.38 & 97.24 & & 51.05 & 83.28 & \textbf{91.80}  \\
\midrule
\ours $(k=1)$ & & \textbf{70.80} & 93.16 & 97.24 & & 50.77 & 82.76 & 91.31 \\

\ours $(k=2)$ & & 70.36  & \textbf{93.46} & 97.20 & & \textbf{51.31} & \textbf{83.38} & 91.69 \\

\ours $(k=3)$ & & 70.42  & 93.34 & 97.22 & & 50.98 & 83.17 & 91.65 \\

\ours $(k=4)$ & & 70.14  & 93.42 & \textbf{97.76} & & 50.82 & 82.66 & 91.14 \\
\bottomrule
\end{tabular}
\end{table}

\bibliographystyle{IEEEtran}
\bibliography{bibliography}

\end{document}